\documentclass[10pt,twocolumn,letterpaper]{article}

\usepackage{iccv}
\usepackage{times}
\usepackage{epsfig}
\usepackage{graphicx}
\usepackage{amsmath}
\usepackage{amssymb}

\usepackage{graphicx}
\usepackage{amsmath}
\usepackage{amssymb}
\usepackage{booktabs}
\usepackage{multirow}
\usepackage{multicol}
\usepackage{makecell}
\usepackage{booktabs}
\usepackage{colortbl}
\usepackage{xcolor}
\usepackage{algorithm} 
\usepackage{algpseudocode} 
\usepackage{pythontex}
\usepackage[accsupp]{axessibility}

\usepackage[pagebackref=true,breaklinks=true,letterpaper=true,colorlinks,bookmarks=false]{hyperref}

\iccvfinalcopy % *** Uncomment this line for the final submission

\def\iccvPaperID{3704} % *** Enter the ICCV Paper ID here

\ificcvfinal\fi

\begin{document}

%%%%%%%%% TITLE

\title{Bridging Cross-task Protocol Inconsistency for Distillation \\ in Dense Object Detection}

\newcommand*\samethanks[1][\value{footnote}]{\footnotemark[#1]}

\author{Longrong Yang$^1$\thanks{The first two authors contributed equally to this paper.} , Xianpan Zhou$^2$\samethanks[1] , Xuewei Li$^1$, Liang Qiao$^{1,3}$ \\ Zheyang Li$^{1,3}$, Ziwei Yang$^3$, Gaoang Wang$^4$, Xi Li$^{1,5,6}$\thanks{Corresponding author.}
\vspace{1mm}
\\
$^1$College of Computer Science \& Technology, Zhejiang University
\\
$^2$Polytechnic Institute, Zhejiang University 
\\$^3$Hikvision Research Institute \enspace
$^4$ZJU – UIUC Institute, Zhejiang University
\\
$^5$Shanghai Institute for Advanced Study of Zhejiang University
\\
$^6$Zhejiang – Singapore Innovation and AI Joint Research Lab, Hangzhou
\\
{\tt \small \{longrongyang, zhouxianpan, xueweili, xilizju\}@zju.edu.cn}
\\
{\tt \small \{qiaoliang6, lizheyang, yangziwei5\}@hikvision.com}, 
{\tt \small gaoangwang@intl.zju.edu.cn}
}

\maketitle
% Remove page # from the first page of camera-ready.
\ificcvfinal\thispagestyle{empty}\fi

%%%%%%%%% ABSTRACT
\begin{abstract}
   
    Knowledge distillation (KD) has shown potential for learning compact models in dense object detection. 
    However, the commonly used softmax-based distillation ignores the absolute classification scores for individual categories. Thus, the optimum of the distillation loss does not necessarily lead to the optimal student classification scores for dense object detectors. This cross-task protocol inconsistency is critical, especially for dense object detectors, since the foreground categories are extremely imbalanced.
    To address the issue of protocol differences between distillation and classification, we propose a novel distillation method with cross-task consistent protocols, tailored for the dense object detection.
    For classification distillation, we address the cross-task protocol inconsistency problem by formulating the classification logit maps in both teacher and student models as multiple binary-classification maps and applying a binary-classification distillation loss to each map. 
    For localization distillation, we design an IoU-based Localization Distillation Loss that is free from specific network structures and can be compared with existing localization distillation losses. 
    Our proposed method is simple but effective, and experimental results demonstrate its superiority over existing methods.
    Code is available at \url{https://github.com/TinyTigerPan/BCKD}.
\end{abstract}

%%%%%%%%% BODY TEXT

\vspace{-0.5cm}
\section{Introduction}

\begin{figure}[tb]
\begin{center}
\graphicspath{{images/}}
\includegraphics[width=1.0\linewidth]{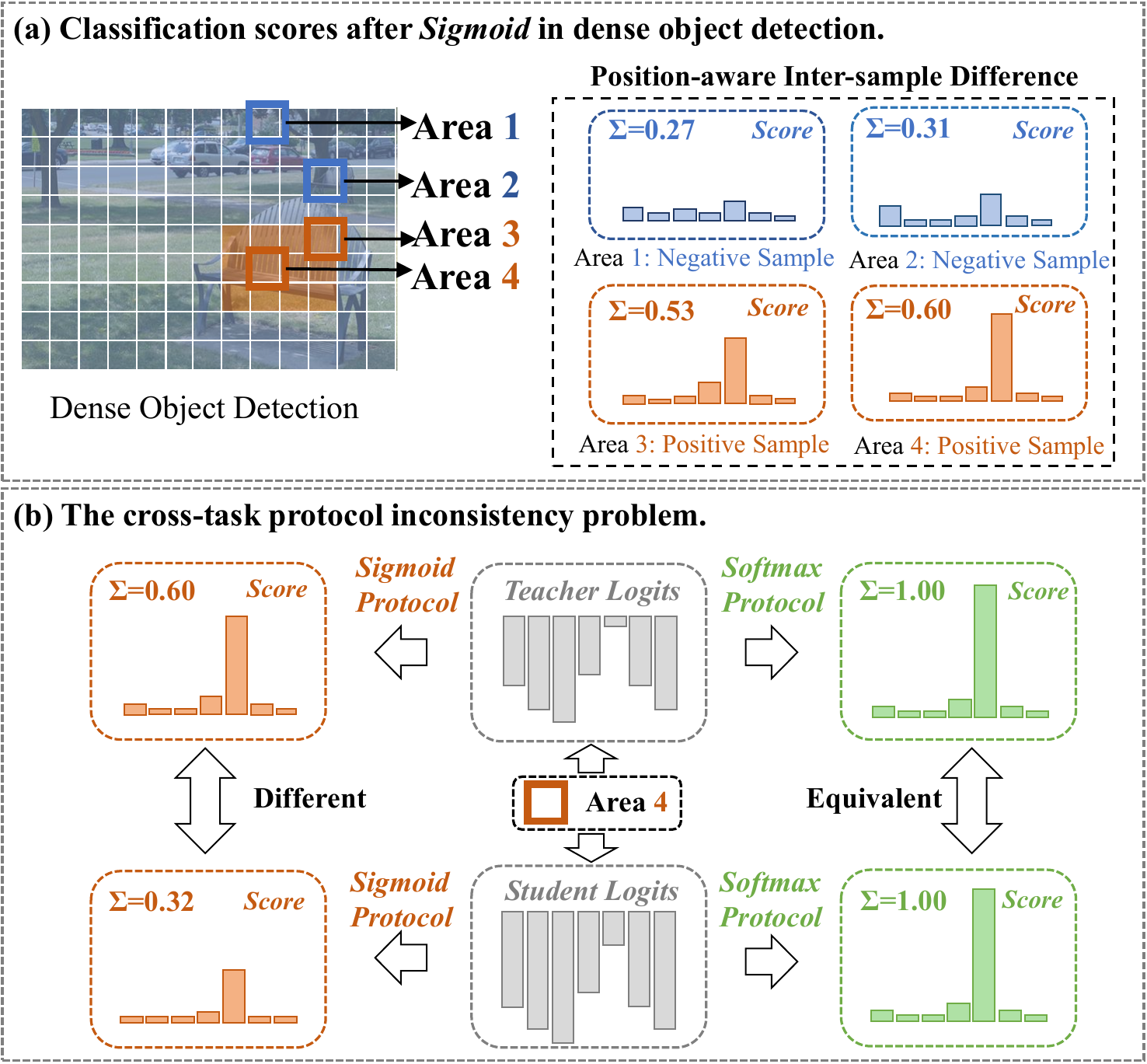}
\end{center}
    \vspace{-4mm}
  \caption{(a) In dense object detection, different samples exhibit inter-sample differences in their classification score sums on various positions on dense maps, which is significantly different from those in image classification.
  (b) The cross-task protocol inconsistency problem arises in dense object detection due to the mismatch between \textit{Sigmoid} protocol used in this task and \textit{Softmax} protocol used in classification distillation. Specifically, when classification distillation loss equals 0, inconsistencies emerge between the scores of the student and teacher models in dense object detection.}
  \vspace{-6mm}
\label{intro}
\end{figure}

Recent progress in dense object detectors has yielded significant performance improvements in the object detection task~\cite{yolo, fcos, retinanet, gfl, wu2023boosting}.
However, the high computational burden of existing detection methods poses a significant challenge for deployment on resource-constrained devices. 
To address this problem, knowledge distillation (KD)~\cite{hinton2015distilling,chen2017learning,defeat,FGD,MGD,PKD,rankmimicking,lad,LD,dai2021general,zhang2022lgd} has emerged as a promising approach to compress models. 
The KD framework involves training a smaller student model by leveraging a larger and more capable teacher model, for enhancing the student model's generalization ability.

Knowledge distillation approaches can be roughly classified into two categories: feature-based distillation methods~\cite{romero2014fitnets,yim2017gift,park2019relational,ahn2019variational, park2019relational,PKD,MGD} and logit-based distillation methods~\cite{hinton2015distilling,zhao2022decoupled,yang2022rethinking,LD}.
In object detection, existing knowledge distillation methods have focused primarily on feature-based distillation due to the marginal performance gain from original logit-based distillation techniques~\cite{kang2021instance,wang2019distilling,zhang2021improve}. 
However, it is worth exploring logit-based methods as they are usually simpler to use and have the potential to further improve performance when combined with feature-based methods.
LD~\cite{LD} is a representative logit-based distillation technique that transforms bounding boxes into probability distributions to facilitate localization distillation.
However, classification distillation in dense object detection remains a challenge.

In this work, we further investigate this problem. 
Figure~\ref{intro} (a) demonstrates that dense object detection faces a severe foreground-background imbalance problem when predicting classification scores on dense maps.
Consequently, dense object detectors typically use the \textit{Sigmoid} protocol to transfer classification logits to classification scores, which results in the position-aware inter-sample difference: Samples closer to positive sample regions generate higher classification score sums across all categories, indicating inter-sample differences. 
However, common classification distillation methods~\cite{hinton2015distilling,zhao2022decoupled,yang2022rethinking,LD} directly use the \textit{Softmax} protocol from image classification to transfer classification logits to classification scores. 
The \textit{Softmax} protocol normalizes classification scores, ignoring the absolute classification scores for individual categories and eliminating the inter-sample difference characteristic of classification scores.
Additionally, in distillation, classification scores for each category are jointly optimized with inter-class dependencies, while in dense object detection, they are individually optimized without such dependencies. 
These differences lead to the {\bf cross-task protocol inconsistency} problem, as shown in Figure~\ref{intro} (b): when the teacher scores are equal to the student scores after \textit{Softmax}, the classification distillation loss is 0, indicating that the student scores have achieved the optimal solution in the distillation loss.
However, after \textit{Sigmoid}, the student scores still differ from the teacher scores, showing lower score sums and incorrect inter-class relationships.

%%%%%%% IoU loss
In addition to classification, localization is another crucial aspect of the object detection task. 
Although the localization distillation loss in LD~\cite{LD} has demonstrated effectiveness, it requires the use of a Discrete Position-probability Prediction Head, such as the Generalized Focal Loss Head~\cite{gfl}, for accurately predicting the localization probability distribution of each sample.
Unfortunately, current object detectors~\cite{yolo,fcos,retinanet} commonly use a Continuous Box-Offset Prediction Head, which means that the use of LD~\cite{LD} would require specific training of teacher models to incorporate the Discrete Position-probability Prediction Head.
This constraint limits the applicability of LD~\cite{LD}.

To address these issues outlined above, this paper proposes two novel distillation losses, Binary Classification Distillation Loss and IoU-based Localization Distillation Loss, tailored for classification and localization in dense object detectors.
For classification, we convert cross-task {\bf inconsistent} protocols into cross-task {\bf consistent} protocols. 
Specifically, we treat the classification logit maps used in dense object detectors as $K$ (\emph{i.e.}, the number of categories) binary-classification maps. 
Then, we use the \textit{Sigmoid} protocol to obtain scores and apply a binary cross entropy loss to distill each binary-classification map from teacher to student models, effectively solving the cross-task protocol inconsistency problem. 
For localization, we convert the special-structure-\textbf{dependent} localization distillation loss into a special-structure-\textbf{free} localization distillation loss.
Specifically, we directly compute the Intersection over Unions (IoUs) between predicted bounding boxes generated by the teacher and student models and employ the IoU loss to minimize the difference between the IoU values and 1 (\emph{i.e.}, the maximal IoU).
Our approach is evaluated on widely used COCO~\cite{coco} dataset, and our experimental results demonstrate that our method outperforms existing logit-based distillation methods and further boosts the existing feature-based distillation methods.
Our contributions are summarized as follows:

$(i)$ We identify the cross-task protocol inconsistency problem as the primary obstacle in utilizing original classification distillation techniques for dense object detection. 
The proposed Binary Classification Distillation Loss greatly enhances the performance gains obtained through classification distillation in dense object detection.
We show that transferring semantic knowledge (\emph{i.e.}, classification) alone can be effective in dense object detection, beyond common views in previous work.

$(ii)$ We propose the IoU-based Localization Distillation Loss to distill the localization knowledge from teacher models to student models, which eliminates the need for specific training of teacher models.

$(iii)$ Our proposed method is simple but effective, as demonstrated by our experiments. 
Besides, our method exhibits flexibility in integrating with existing state-of-the-art methods, resulting in a consistent performance increase.

\vspace{-1mm}
\section{Related Works}

% %-------------------------------------------------------------------------
\subsection{Object Detection}

Object detection is a fundamental and challenging task in computer vision, involving the classification and localization of objects within a given image.
The literature on this topic can be broadly classified into two categories: region-based object detectors and dense object detectors. 
Region-based object detectors, including Faster-RCNN~\cite{ren2015faster}, Cascade R-CNN~\cite{cai2018cascade}, and Fast R-CNN~\cite{girshick2015fast}, utilize a Region Proposal Network (RPN) to generate Regions of Interest (RoIs), which are then refined through classification and regression heads to produce the final detection. 
In contrast, dense object detectors, such as YOLO~\cite{yolo}, FCOS~\cite{fcos}, RetinaNet~\cite{retinanet}, and GFL~\cite{gfl}, directly predict objects from feature maps, offering advantages in terms of computational efficiency and ease of deployment when compared to region-based object detectors.

Most dense object detectors generate predictions of various sizes and proportions by utilizing dense proposals (such as anchor~\cite{retinanet} and point~\cite{fcos}) at all positions on the image.
Thus, they face the challenge of a severe imbalance between positive and negative samples, which can lead to poor performance. 
To address this, some works~\cite{liu2016ssd, zhang2018single} have explored complex re-sampling schemes for hard example mining. Besides, RetinaNet~\cite{retinanet} uses the focal loss to prioritize the training of difficult samples. 
Additionally, different label assignment strategies, such as ATSS~\cite{zhang2020atss} and OTA~\cite{ge2021ota}, have been proposed to further improve performance. 
Through collective efforts, dense object detectors have achieved high accuracy and fast inference times. 
Recent research has also focused on improving the performance of compact real-time models through model compression techniques. 
For example, successful approaches include RTMDet~\cite{lyu2022rtmdet} and YOLOv7~\cite{wang2022yolov7}.

\subsection{Knowledge Distillation}

Knowledge Distillation (KD) is a model compression method that enables training of compact student models with guidance from more powerful teacher models. First introduced by Hinton et al. \cite{hinton2015distilling}, KD has since been extensively studied in subsequent works~\cite{romero2014fitnets,yim2017gift,ahn2019variational,phuong2019towards,stanton2021does,zhao2022decoupled, li2021reskd,zhou2021towards,zhou2022modeling,wang2018progressive,fu2021elastic,zhao2021memory,zhao2021mgsvf,zhao2022rbc,summaira2021recent,chen2014ranking,jiang2019learning,li2011graph}.
In classification, KD methods are typically classified into two categories: feature-based methods~\cite{romero2014fitnets,yim2017gift,park2019relational,ahn2019variational} and logits-based methods~\cite{hinton2015distilling,zhao2022decoupled,yang2022rethinking}.  
Feature-based methods transfer knowledge by mimicking intermediate features from a teacher's hint layer, while logits-based methods by mimicking the logit outputs from the teacher's classifier. 
In object detection, KD was initially applied in~\cite{chen2017learning}, and many subsequent works have been proposed \cite{chen2017learning,defeat,FGD,MGD,PKD,rankmimicking,lad,LD,dai2021general,zhang2022lgd} to improve student performance.
Feature-based distillation remains the mainstream approach.
For example, FGD~\cite{FGD} separates foreground and background and recovers missing information by rebuilding relationships among different pixels. 
PKD~\cite{PKD} relaxes constraints on the magnitude of features by mimicking the Pearson Correlation Coefficient.
MGD~\cite{MGD} randomly masks some pixels of the student’s feature and leverages a simple generative block to force it to imitate the teacher’s feature.
DIC~\cite{guo2021distilling} explores the classifier-to-detector knowledge transfer.
TLLM~\cite{zhu2022teach} explores ``undistillable classes", focusing on scenarios where a significant disparity exists between teacher and student.
Regarding logit-based distillation methods, LD~\cite{LD} treats bounding box regression as probability distribution estimation, and argues that distilling localization knowledge is more effective than semantic knowledge in dense object detection.

In previous works, logit-based distillation methods in image classification are directly utilized to distill the semantic knowledge from teacher models to student models, and they commonly find that the semantic knowledge transfer seldom works for dense object detection. 
In this work, we argue that these approaches overlook the differences between object detection and image classification tasks, which leads to insufficient performance gains. 
To address this issue, we propose a novel classification distillation method tailored for dense object detection in this paper.

%%%%%%%

\vspace{-1mm}
\section{Methodology}

\subsection{Overview}

\begin{figure*}[tb]
\begin{center}
\graphicspath{{images/}}
\includegraphics[width=0.85\linewidth]{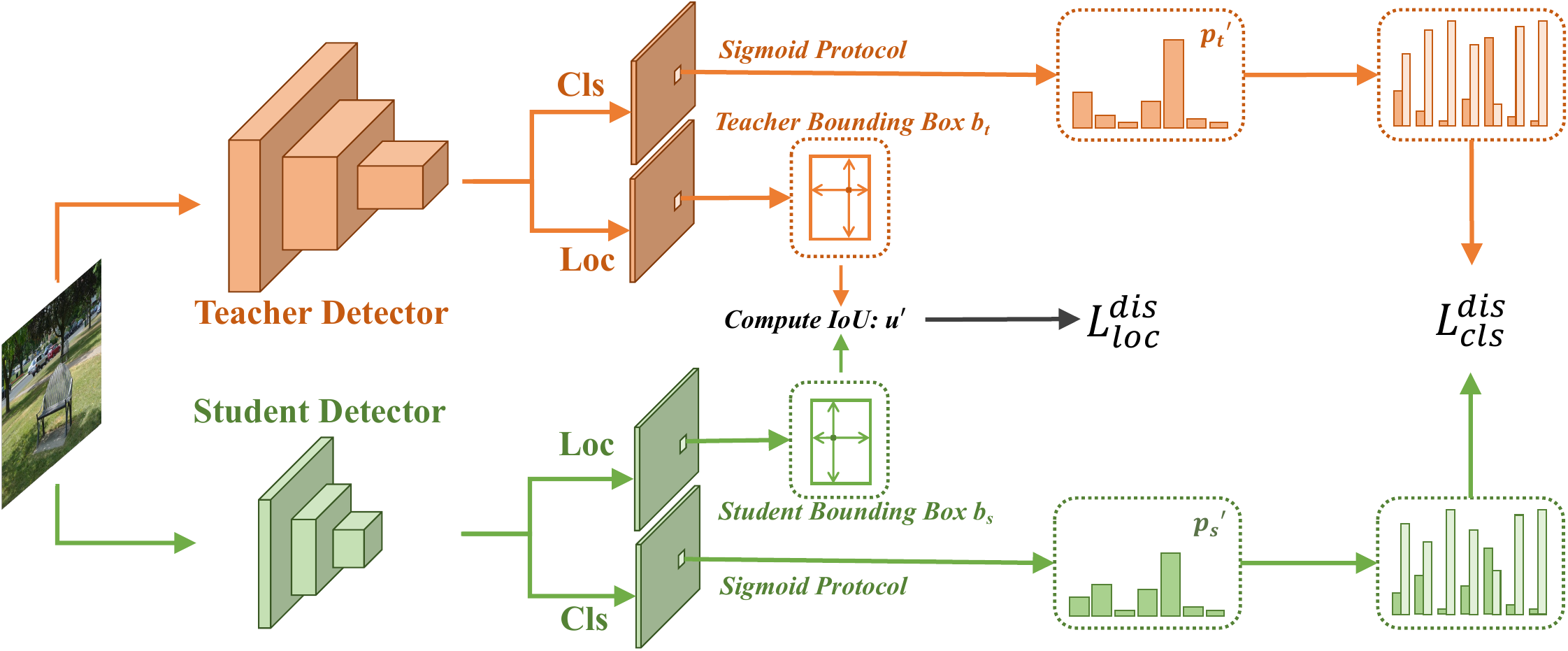}
\end{center}
\vspace{-2mm}
  \caption{Distillation pipeline of our method. 
  We leverage two novel distillation losses tailored for the object detection task. 
  $(i)$ Binary Classification Distillation Loss $\mathcal{L}_{cls}^{dis}$, which represents classification logit maps as multiple binary-classification maps and distills classification knowledge through a distillation loss similar to binary cross entropy. 
  $(ii)$ IoU-based Localization Distillation Loss $\mathcal{L}_{loc}^{dis}$, which transfers localization knowledge from teacher models to student models by computing the IoUs between predicted bounding boxes from both models and using the IoU loss. 
  Best viewed in color.}
\vspace{-2mm}
\label{metho}
\end{figure*}

A dense object detector can be represented as the combination of a feature extractor $f(\cdot)$ and a detection head $h(\cdot)$. 
Given an input image $I$, the detector first extracts features $F$$=$$f(I)$, and then generates the final prediction $P$$=$$h(F)$.
The prediction $P$ typically comprises classification logits $l \in \mathbb{R}^{n \times K}$ and localization offsets $o \in \mathbb{R}^{n \times 4}$, where $n$ is the number of anchors or points in dense object detection, and $K$ is the number of foreground categories.
In existing knowledge distillation (KD) methods for dense object detection, knowledge is transferred from a frozen large teacher detector $T_{det}$ to a small student detector $S_{det}$. 
For feature-based methods, the distillation loss is defined as $\mathcal{L}_{dis} = \text{loss}(F_t, F_s)$, where $F_t$ and $F_s$ indicate the features of $T_{det}$ and $S_{det}$, respectively. 
For logits-based methods, the distillation loss is defined as $\mathcal{L}_{dis} = \text{loss}(P_t, P_s)$, where $P_t$ and $P_s$ indicate the predictions of $T_{det}$ and $S_{det}$, respectively, and $\text{loss}$ denotes the distillation loss function.

In this work, we propose two distillation losses tailored for classification and localization in dense object detection, as illustrated in Figure~\ref{metho}. 
We observe the cross-task protocol inconsistency problem between dense object detection and classification distillation loss, which impedes the effectiveness of the classification distillation in dense object detection. 
To address this problem, we introduce a novel Binary Classification Distillation Loss that converts the inconsistent cross-task protocol distillation into the consistent cross-task protocol distillation.
Moreover, we find that existing localization distillation methods rely on the Discrete Position-probability Prediction Head, such as the Generalized Focal Loss Head~\cite{gfl}, which requires specific training of teacher models. 
To overcome this limitation, we propose a special structure-free IoU distillation loss that enables the distillation of localization knowledge from teacher models to student models.

\subsection{Binary Classification Distillation Loss}
{\bf Protocol in Dense Object Detectors}: Dense object detectors aim to predict the corresponding classification score and bounding box for each sample point in dense maps generated from the entire image. 
However, as the background pixels occupy a significant portion of the image, foreground and background samples are severely imbalanced in dense object detectors. 
Specifically, during training, the majority of the samples are background samples.
When using the \textit{Softmax} protocol for transferring classification logits to classification scores, which assigns a sample to $K$+1 probabilities (where $K$ is the number of foreground categories and an additional probability indicates the background), it may not be effective due to its tendency to assign higher probabilities to the majority class, \emph{i.e.}, the background.
Consequently, dense object detectors such as YOLO~\cite{yolo}, FCOS~\cite{fcos}, RetinaNet~\cite{retinanet}, and GFL~\cite{gfl} commonly use the \textit{Sigmoid} protocol for transferring classification logits to classification scores. 
By modeling the multi-classification problem as multiple binary-classification problems, this approach can more effectively handle the foreground-background class imbalance issue.

Specifically, dense object detectors produce classification maps of varying sizes, with a size of $H$$\times$$W$$\times$$K$, where $H$, $W$ and $K$ represent the height, width and number of classes, respectively. 
Existing methods assign labels to each point on the classification map, with positive samples labeled as a one-hot tensor and negative samples labeled as a fully-zero tensor.
Let $x$ be a sample, and $l$$\in$$\mathbb{R}^{n \times K}$ denote its classification logits. 
To obtain classification scores for each point, existing methods use the \textit{Sigmoid} protocol, \emph{i.e.}, $p=Prot_{Sig}(l)$. 
% $p = \{p_1, p_2, ..., p_K\}$. 
We also have a label tensor $y$ for $x$. 
Therefore, we can compute the binary cross entropy loss between the classification scores and labels:
\begin{equation}
\mathcal{L}_{cls}(x) = \sum_{i=1}^{n} \sum_{j=1}^{K} \mathcal{L}_{CE}(p_{i,j}, y_{i,j}),
\end{equation}
where $\mathcal{L}_{CE}(p_{i,j},y_{i,j})$ is the binary cross entropy loss for the $i$-th position and $j$-th class, defined as:
\begin{equation}
\mathcal{L}_{CE}(p_{i,j},y_{i,j}) = \left\{
\begin{aligned}
& -log(p_{i,j}) \quad &y_{.,j}=1, \\
& -log(1-p_{i,j}) \quad &y_{.,j}=0.
\end{aligned}
\right.
\end{equation}

{\bf Protocol in Common Classification Distillation}: Common classification distillation methods~\cite{kang2021instance,wang2019distilling,zhang2021improve,LD}  are usually developed for the class-balanced scenario in image classification. 
The \textit{Softmax} protocol plays a crucial role in establishing strong inter-class relationships, providing strong discriminative ability for identifying different categories in image classification. 
Therefore, the \textit{Softmax} protocol is typically used in classification distillation.

Specifically, for a sample $x$, let $l^t$ and $l^s$ denote the classification logits from the teacher and student models, respectively. 
Existing methods use the \textit{Softmax} protocol to obtain classification scores, \emph{i.e.}, $p^t$$=$$Prot_{Smax}(l^t)$ and $p^s$$=$$Prot_{Smax}(l^s)$.
The classification distillation loss is computed between $p^t$ and $p^s$ to encourage the student model to mimic the output of the teacher model. 
Specifically, this loss is typically defined as the Kullback-Leibler (KL) divergence between teacher scores and student scores:
\begin{equation}
\mathcal{L}_{cls}^{kl}(x) = \mathcal{L}_{kl}(p^s, p^t),
\end{equation}
where $\mathcal{L}_{cls}^{kl}(\cdot)$ denotes the classification distillation loss, and $\mathcal{L}_{kl}(\cdot, \cdot)$ denotes the Kullback-Leibler (KL) divergence.

\vspace{0.1cm}
{\bf Analysis of Cross-task Protocol Inconsistency}: 
Existing distillation methods~\cite{LD} in object detection typically apply the classification distillation loss used in image classification directly to dense object detection, leading to cross-task protocol inconsistency. 
Specifically, we firstly present the \textit{Softmax} protocol and the \textit{Sigmoid} protocol below:
\begin{equation}
Prot_{Smax}(l^t) = \frac{e^{l^t}}{\sum_{i=1}^{K}e^{l_{i}^{t}}}, Prot_{Sig}(l^t) = \frac{1}{1+e^{-l^t}},
\end{equation}
where $l^t$ is the logits of the teacher model. When $n$ is a constant tensor with the same shape with $l^t$ and $l^s = l^t + n$ ($l^s$ is the logits of the student model), we have:
\begin{equation}
\begin{aligned}
Prot_{Smax}(l^s) &= \frac{e^{l^t+n}}{\sum_{i=1}^{K}e^{l_{i}^{t}+n}} = \frac{e^{l^t} \cdot e^{n}}{\sum_{i=1}^{K}(e^{l_{i}^{t}} \cdot e^{n})} \\
&= \frac{e^{l^t}}{\sum_{i=1}^{K}e^{l_{i}^{t}}} = Prot_{Smax}(l^t).
\end{aligned}
\end{equation}
Thus, the distillation loss is equal to zero, and there is no further transfer of localization knowledge from the teacher to the student model. However, $Prot_{Sig}(l_s) \ne Prot_{Sig}(l_t)$, resulting in a significant gap between the classification scores of the teacher and student models during inference. Typically, the scores obtained by the student model are lower than those of the teacher model and may have incorrect inter-class relationships. As a result, the student model cannot inherit the correct prediction ability from the teacher model. 

{\bf Bridge Cross-task Protocol Inconsistency}: To bridge cross-task protocol inconsistency, we propose a straightforward but effective solution. 
Specifically, we treat classification logit maps as multiple binary-classification maps during distillation. 
To achieve this, we compute ${p^t}' = Prot_{Sig}(l^{t})$ and ${p^s}' = Prot_{Sig}(l^{s})$, resulting in the binary-classification scores ${p^t}'$ and ${p^s}'$ with a size of $n \times K$. 
The classification distillation loss can then be calculated based on these binary-classification scores:
\begin{equation}
% \begin{small}
\begin{aligned}
& \mathcal{L}_{BCE}({p_{i,j}^s}',{p_{i,j}^t}') = \\
& - ((1 - {p_{i,j}^t}') \cdot log(1 - {p_{i,j}^s}') + {p_{i,j}^t}' \cdot log({p_{i,j}^s}')), \\
& \mathcal{L}_{cls}^{dis}(x) = \sum_{i=1}^{n} \sum_{j=1}^{K} \mathcal{L}_{BCE}({p_{i,j}^s}',{p_{i,j}^t}'),
\end{aligned}
% \end{small}
\end{equation}
where $\mathcal{L}_{cls}^{dis}(\cdot)$ denotes the classification distillation loss, and $\mathcal{L}_{BCE}(\cdot,\cdot)$ denotes the binary cross entropy loss, ${p_{i,j}^{s}}'$, ${p_{i,j}^{s}}'$ denotes the $i$-th position and $j$-th class of ${p^{s}}'$, ${p^{t}}'$, respectively.

Besides, we propose a loss weighting strategy for models to focus on distilling important samples, inspired by the Focal Loss~\cite{retinanet}. Specifically, we compute the importance weighting $w$ of the sample $x$ as follows:
% \begin{equation}
% w = \left| \mathop{max}(p_{.,j}^t') - \mathop{max}(p_{.,j}^s') \right|,
% \end{equation}

\begin{equation}
w = \left| {p^t}' - {p^s}' \right|,
\end{equation}
where $w \in \mathbb{R}^{n \times K}$. 
Each element in $w$ weighted to the classification distillation loss of sample $x$. 
Thus, the classification distillation loss in this paper is formulated as:
\begin{equation}
\mathcal{L}_{cls}^{dis}(x) = \sum_{i=1}^{n} \sum_{j=1}^{K} w_{i,j} \cdot \mathcal{L}_{BCE}({p_{i,j}^{s}}',{p_{i,j}^{t}}').
\end{equation}

\subsection{IoU-based Localization Distillation Loss}
In addition to classification, another crucial aspect of object detection is localization. 
LD~\cite{LD} transforms the bounding box into a probability distribution to tackle the localization distillation problem. 
In LD~\cite{LD}, a Discrete Position-Probability Prediction Head, such as the Generalized Focal Loss Head~\cite{gfl}, is essential for precisely predicting the localization probability distribution of each sample.
Regrettably, this type of head is not commonly employed in current object detectors~\cite{yolo,fcos,retinanet} due to their complexity, especially in inference, resulting in the need for specific training of teacher models.
To address this issue, we propose an innovative structure-free localization distillation loss, motivated by the Interaction-over-Union (IoU) loss widely used in dense object detectors, to replace the existing ones.

\begin{table*}[!ht]
  \begin{center}
  
  \begin{tabular}{l|c|cccccc}
    \toprule
    Method & Schedule & mAP & AP$_{50}$ & AP$_{75}$ & AP$_{S}$ & AP$_{M}$ & AP$_{L}$\\
    \midrule
    \rowcolor{lightgray!45}GFocal-Res101(Teacher)  & 2x & 44.9 & 63.1 & 49.0 & 28.0 & 49.1 & 57.2\\
    \rowcolor{lightgray!45}GFocal-Res50(Student)   & 1x & 40.1 & 58.2 & 43.1 & 23.3 & 44.4 & 52.5\\
    LD~\cite{LD}                 & 1x & 42.1(+2.0) & 60.3(+2.1) & 45.6(+2.5) & 24.5(+1.2) & 46.2(+1.8) & 54.8(+2.3)\\

    Ours      & 1x & \textbf{43.2(+3.1)} & \textbf{61.6(+3.4)} & \textbf{46.9(+3.8)} & \textbf{25.7(+2.4)} & \textbf{47.3(+2.9)} & 55.9(+3.4) \\
    LD~\cite{LD} + Ours & 1x & \textbf{43.2(+3.1)} & 61.4(+3.2) & 46.7(+3.6) & 25.1(+1.8) & \textbf{47.3(+2.9)} & \textbf{56.1(+3.6)} \\

    \midrule
    \rowcolor{lightgray!45}GFocal-Res101(Teacher)  & 2x & 44.9 & 63.1 & 49.0 & 28.0 & 49.1 & 57.2\\
    \rowcolor{lightgray!45}GFocal-Res34(Student)   & 1x & 38.9 & 56.6 & 42.2 & 21.5 & 42.8 & 51.4\\
    LD~\cite{LD}                 & 1x & 41.0(+2.1) & 58.6(+2.0) & 44.6(+2.4) & 23.2(+1.7) & 45.0(+2.2) & 54.2(+2.8)\\
    Ours      & 1x & 42.0(+3.1) & 60.0(+3.4) & 45.6(+3.4) & 24.1(+2.6) & 46.3(+3.5) & 54.1(+2.7) \\
    LD~\cite{LD} + Ours & 1x & \textbf{42.3(+3.4)} & \textbf{60.2(+3.6)} & \textbf{46.0(+3.8)} & \textbf{24.4(+2.9)} & \textbf{46.4(+3.6)} & \textbf{54.8(+3.4)} \\

    \midrule
    \rowcolor{lightgray!45}GFocal-Res101(Teacher)  & 2x & 44.9 & 63.1 & 49.0 & 28.0 & 49.1 & 57.2\\
    \rowcolor{lightgray!45}GFocal-Res18(Student)   & 1x & 35.8 & 53.1 & 38.2 & 18.9 & 38.9 & 47.9\\
    LD~\cite{LD}     & 1x & 37.5(+1.7) & 54.7(+1.6) & 40.4(+2.2) & 20.2(+1.3) & 41.2(+2.3) & 49.4(+1.5)\\
    Ours      & 1x & 38.6(+2.8) & 56.4(+3.3) & 41.7(+3.5) & 21.4(+2.5) & 42.0(+3.1) & 50.0(+2.1) \\
    LD~\cite{LD} + Ours & 1x & \textbf{38.9(+3.1)} & \textbf{56.6(+3.5)} & \textbf{42.0(+3.8)} & \textbf{22.2(+3.3)} & \textbf{42.5(+3.6)} & \textbf{50.8(+2.9)} \\
    
    \bottomrule
  \end{tabular}
  \end{center}
  \caption{Quantitative results of the proposed method and existing logits-based distillation methods for lightweight detectors. All results are evaluated on MS COCO \textit{val2017}. Boldface indicates the best results.}
  \vspace{-2mm}
  \label{table:main results}
\end{table*}

LD~\cite{LD} discretizes the continuous regression range into a uniform discrete variable $[e_1,e_2,...,e_n]^T$ with $n$ intervals. 
To predict the $n$ logits corresponding to each regression interval of each edge $e$, denoted by $z_T$ and $z_S$ for the teacher and student, respectively, a Discrete Position-Probability Prediction Head (\emph{e.g.}, the Generalized Focal Loss Head) is needed. 
The generalized \textit{Softmax} function is then employed to transform $z_T$ and $z_S$ into the probability distribution $p_T$ and $p_S$, respectively. 
Finally, the Kullback-Leibler Divergence is used to minimize the distance between $p_T$ and $p_S$.
Although effective, this approach requires the use of a specific head, namely the Generalized Focal Loss Head, to predict discrete logits for all possible positions of each edge. 
Instead, these detectors typically predict continuous bounding box offsets that are more convenient for obtaining the predicted bounding box in inference. 
Therefore, the applicability of LD~\cite{LD} is limited.

In this work, our objective is to transfer localization knowledge from teacher models to student models without relying on complex transformations of bounding box predictions. 
To achieve this, we leverage the most fundamental location relationship between two bounding boxes, Intersection over Union (IoU), as the distillation target. 
Specifically, we obtain localization maps from both the teacher and student models, and for a given input sample $x$, we denote the corresponding localization predictions from the teacher and student models in $i$-th position as $o_{i}^t$ and $o_{i}^s$, respectively. 
We then obtain the bounding box for $x$ by using the anchor position and localization prediction, where $A_{i}$ denotes the $i$-th anchor. 
The bounding box for the teacher model and student model are obtained as $b_{i}^t=Decoder(A_{i},o_{i}^t)$ and $b_{i}^s=Decoder(A_{i},o_{i}^s)$, respectively.
We compute the IoU between $b_{i}^t$ and $b_{i}^s$, denoted as $u_{i}'$. 
In addition, we introduce a loss weighting strategy for models to focus on distilling important samples in the above section, which we also use for the localization distillation. 
Therefore, the localization distillation loss can be computed as:
\begin{equation}
\mathcal{L}_{loc}^{dis}(x) = \sum_{i=1}^{n} \mathop{max}(w_{.,j}) \cdot (1 - u_{i}').
\end{equation}
The localization distillation loss is straightaway but comparable to existing localization distillation losses.

\subsection{Total Distillation Loss}
In this work, we introduce two novel distillation losses, namely Binary Classification Distillation Loss and IoU-based Localization Distillation Loss, for improving the performance of both classification and localization tasks. 
The proposed classification distillation loss is specifically designed for the classification task, whereas the IoU loss is developed for the localization task. 
The combined distillation loss is formulated as follows:
\begin{equation} 
\label{loss_total}
\mathcal{L}_{total}^{dis}(x) = \alpha_{1} \cdot \mathcal{L}_{cls}^{dis}(x) + \alpha_{2} \cdot \mathcal{L}_{loc}^{dis}(x),
\end{equation}
where $\alpha_{1}$ and $\alpha_{2}$ are two hyper-parameters, denoting the loss weightings for the classification distillation loss and the localization distillation loss, respectively.

\begin{table*}[!ht]
  \begin{center}
  
  \begin{tabular}{l|c|cccccc}
    \toprule
    Method & Schedule & mAP & AP$_{50}$ & AP$_{75}$ & AP$_{S}$ & AP$_{M}$ & AP$_{L}$\\
    \midrule
    
    \rowcolor{lightgray!45}RetinaNet-ResX101(Teacher) & 1x & 41.0 & 60.9 & 43.9 & 23.9 & 45.2 & 54.0\\
    \rowcolor{lightgray!45}RetinaNet-Res50(Student)   & 1x & 36.5 & 55.4 & 39.1 & 20.4 & 40.3 & 48.1\\
    Ours                  & 1x & 39.2(+2.7) & 58.7(+3.3) & 42.0(+2.9) & 22.4(+2.0) & 43.1(+2.8) & 52.1(+4.0)\\
    
    MGD \cite{MGD}         & 1x & 39.6(+3.1) & 59.0(+3.6) & 42.4(+3.3) & \textbf{22.7(+2.3)} & 43.9(+3.6) & 53.0(+4.9)\\
    PKD \cite{PKD}         & 1x & 39.7(+3.2) & 59.0(+3.6) & 42.4(+3.3) & 22.5(+2.1) & 44.2(+3.9) & 53.7(+5.6)\\
    
    MGD \cite{MGD} + Ours  & 1x & \textbf{40.1(+3.6)} & 59.5(+4.1) & \textbf{43.0(+3.9)} & 22.3(+1.9) & \textbf{44.3(+4.0)} & 53.3(+5.2)\\
    PKD \cite{PKD} + Ours  & 1x & \textbf{40.1(+3.6)} & \textbf{59.6(+4.2)} & 42.8(+3.7) & 22.3(+1.9) & \textbf{44.3(+4.0)} & \textbf{53.8(+5.7)}\\
    \midrule

    \rowcolor{lightgray!45}FCOS-Res101(Teacher) & 2x & 40.8 & 60.0 & 44.0 & 24.2 & 44.3 & 52.4\\
    \rowcolor{lightgray!45}FCOS-Res50(Student)  & 1x & 36.6 & 56.0 & 38.8 & 21.0 & 40.6 & 47.0\\
    Ours                  & 1x & 39.2(+2.6) & 58.8(+2.8) & 42.0(+3.3) & 22.7(+1.7) & 43.2(+2.6) & 50.3(+3.3)\\
    
    MGD \cite{MGD}         & 1x & 39.6(+3.0) & 59.0(+3.0) & 42.3(+3.5) & 23.1(+2.1) & 43.7(+3.1) & 51.1(+4.1)\\
    PKD \cite{PKD}         & 1x & 39.9(+3.3) & 59.3(+3.3) & 42.6(+3.8) & 22.9(+1.9) & 44.3(+3.7) & \textbf{51.4(+4.4)}\\
    
    MGD \cite{MGD} + Ours  & 1x & 40.0(+3.4) & 59.3(+3.3) & 42.9(+4.1) & 23.4(+2.4) & 44.1(+3.5) & 51.1(+4.1)\\
    PKD \cite{PKD} + Ours  & 1x & \textbf{40.2(+3.6)} & \textbf{59.5(+3.5)} & \textbf{43.0(+4.2)} & \textbf{23.7(+2.7)} & \textbf{44.5(+3.9)} & \textbf{51.4(+4.4)}\\
    \bottomrule
  \end{tabular}
  \end{center}
  \caption{Quantitative results of the proposed method combined with existing feature-based methods on different dense object detectors. All results are
evaluated on MS COCO \textit{val2017}. Boldface indicates the best results.}
  \vspace{-2mm}
  \label{table:feature-based}
\end{table*}
\section{Experimental and Results}

\subsection{Datasets and Evaluation Metrics}

To verify the effectiveness of the proposed method, we conducted experiments on the popular MS COCO dataset \cite{coco}, which contains about 118k images in the \textit{train} set, 5k in the \textit{val} set, and 20k in the \textit{test-dev} set spanning 80 categories. 
We choose the \textit{train} set for training and the \textit{val} set for testing. 
We report the detection mean average precision (mAP) as an evaluation metric, meanwhile under the different thresholds (\emph{e.g.} AP$_{50}$) and scales (\emph{e.g.} AP$_{S}$).

\subsection{Main Results}
In this paper, we rethink the limitations of the original Knowledge Distillation (KD) approach in dense object detection, and propose two novel distillation losses, namely the Binary Classification Distillation Loss and the IoU-based Localization Distillation Loss, to address the shortcomings of KD in the context of both classification (\textbf{Cls}) and localization (\textbf{Loc}) in detectors. 
Our proposed approach achieves notable performance improvements over the baseline method, without any additional costs.

Our proposed approach yields notable object detection performance improvements, as shown in Table~\ref{table:main results}.
Specifically, we achieve mAP score improvements of +2.8, +3.1, and +3.1 when using GFocal-Res18, GFocal-Res34, and GFocal-Res50 as student models, respectively, significantly outperforming the state-of-the-art method LD~\cite{LD}.
 Moreover, we achieve further mAP score improvements of +0.3 when combining LD~\cite{LD} with our proposed method in GFocal-Res18 and GFocal-Res34.

Feature-based distillation methods such as MGD~\cite{MGD} and PKD~\cite{PKD} have shown powerful performance improvements.
Fortunately, our method can be easily combined with these approaches to further enhance detector performance. 
As illustrated in Table~\ref{table:feature-based}, our method achieves mAP score improvements of +0.4 and +0.5 over PKD and MGD, respectively, when using RetinaNet as the basic dense object detector.
Moreover, our proposed method is highly flexible and can be used with various dense object detectors. 
In the case of FCOS, our approach leads to significant performance improvements. 
Similar to the results with RetinaNet, our method yields mAP score improvements of +0.3 and +0.4 over PKD and MGD, respectively.

\subsection{Ablation Analysis}

\begin{table}[h]
  \begin{center}
  \begin{tabular}{@{}c|cc|ccc}
    \toprule
    Method & \multicolumn{5}{c}{GFocal Res101-Res50}\\
    \midrule
     & \textit{Cls} & \textit{Loc} & mAP & AP$_{50}$ & AP$_{75}$\\
    \midrule
    Baseline & & & 40.1 & 58.2 & 43.1\\
    \midrule
    \multirow{3}{*}{LD~\cite{LD}} & \checkmark &            & 40.4 & 58.9 & 43.4\\
                                  &            & \checkmark & 41.8 & 59.5 & 45.4\\
                                  & \checkmark & \checkmark & 42.1 & 60.3 & 45.6\\
    \midrule
    \multirow{3}{*}{Ours} & \checkmark &            & 42.0 & 60.9 & 45.6\\
                          &            & \checkmark & 42.3 & 60.0 & 45.9\\
                          & \checkmark & \checkmark & \textbf{43.2} & \textbf{61.6} & \textbf{46.9}\\
    \bottomrule
  \end{tabular}
  \end{center}
  \caption{Ablation study of distillation losses on different branch in detectors. \textit{Cls} and \textit{Loc} indicates distillation on classification and localization in detector head, respectively. which are represented as  $\mathcal{L}_{cls}^{dis}(x)$ and $\mathcal{L}_{loc}^{dis}(x)$ in our proposed method. Boldface indicates the best results.}
  \label{table:ablation}
\end{table}
\noindent \textbf{Sensitivity Study of Different Losses.}
To demonstrate the effectiveness of our proposed Binary Classification Distillation Loss ($\mathcal{L}_{cls}^{dis}(x)$) and IoU-based Localization Distillation Loss ($\mathcal{L}_{loc}^{dis}(x)$), we conduct experiments on the GFocal student model. 
As shown in Table~\ref{table:ablation}, both $\mathcal{L}_{cls}^{dis}(x)$ and $\mathcal{L}_{loc}^{dis}(x)$ contribute to improved detector performance, particularly in AP$_{50}$ and AP$_{75}$, which more impacts classification and localization performance, respectively. 
Furthermore, the combination of the two losses leads to significant performance improvements compared to the baseline.

\begin{figure*}[!t]
  \centering
  \graphicspath{{images/}}
  \includegraphics[width=1.0\linewidth]{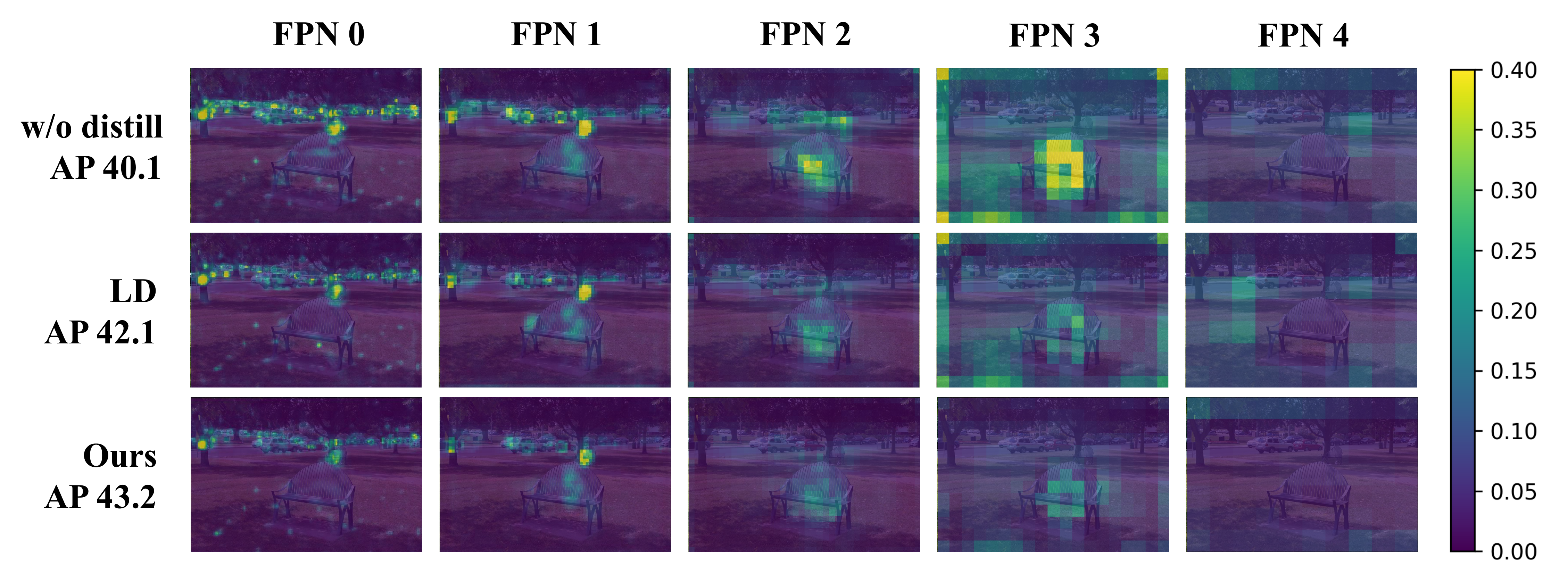}
  \vspace{-6mm}
  \caption{Visualization of L1 error summation of the classification score after \textit{Sigmoid} between the teacher (GFocal-Res101) and the student (GFocal-Res50) at different levels of the Feature Pyramid Network (FPN).
  We can observe that our proposed method achieves a significant reduction in errors for almost all locations compared to the state-of-the-art method LD~\cite{LD}.
  To better observe subtle differences, we bound the margin of error between 0 and 0.4.
  Darker is better. 
  Best viewed in color.
  }
  \vspace{-2mm}
  \label{logits_diff}
\end{figure*}

\vspace{0.1cm}
\noindent \textbf{Sensitivity Study of Different Hyper-parameters.}
Our proposed method employs two hyper-parameters, $\alpha_{1}$ and $\alpha_{2}$, to balance the Binary Classification Distillation Loss and IoU-based Localization Distillation Loss.
As shown in Table~\ref{table:aplha1ablation} and Table~\ref{table:aplha2ablation}, the experiments demonstrate that our method is insensitive to the hyper-parameters and various values of $\alpha_{1}$ and $\alpha_{2}$ can lead to similar significant improvements in performance.
Besides, we can achieve the best quantitative results when setting $\alpha_{1}=1.0$ and $\alpha_{2}=4.0$.

\vspace{0.1cm}
\noindent \textbf{Visualization.}
In order to demonstrate the effectiveness of our proposed method in reducing classification errors, we compared the performance of the teacher detector and the student detector by forwarding the same image to both and recording the L1 error of summation of the classification score after \textit{Sigmoid}.
Figure~\ref{logits_diff} shows that our proposed method significantly reduces the ambiguity in classifying teachers and students in almost all locations at all FPN levels, thus validating the effectiveness of our method.

\begin{table}[h]
  \begin{center}
  \begin{tabular}{@{}c|c|cccccc}
    \toprule
    $\alpha_{1}$ & 0 & 0.25 & 0.5 & 1.0 & 1.5 & 2.0 & 3.0\\
    \midrule
    mAP       & 40.1 & 41.5 & 41.9 & \textbf{42.0} & 41.9 & 41.5 & 41.2\\
    AP$_{50}$ & 58.2 & 60.1 & 60.8 & \textbf{60.9} & 60.6 & 60.6 & 60.2\\
    AP$_{75}$ & 43.1 & 44.9 & 45.4 & \textbf{45.6} & 45.4 & 44.7 & 44.5\\
    % AP$_{S}$  & 23.3 & 24.3 & 24.2 & \textbf{24.6} & 24.5 & \textbf{24.6} & 23.5\\
    % AP$_{M}$  & 44.4 & 45.7 & 46.1 & \textbf{46.2} & 46.1 & 45.6 & 45.5\\
    % AP$_{L}$  & 52.5 & 53.7 & 54.4 & 54.0 & \textbf{54.5} & 53.5 & 53.6\\
    \bottomrule
  \end{tabular}
  \end{center}
  \caption{Ablation study of hyper-parameter $\alpha_{1}$ on GFocal Res101-Res50. To show the sensitivity of $\mathcal{L}_{cls}^{dis}(x)$, we fix $\alpha_{2}=0$. Boldface indicates the best results.}
  \vspace{-2mm}
  \label{table:aplha1ablation}
\end{table}

\begin{table}[h]
  \begin{center}
  \begin{tabular}{@{}c|c|ccccccccc}
    \toprule
    $\alpha_{2}$ & 0 & 0.5 & 1.0 & 1.5 & 2.0 & 4.0 & 5.0\\
    \midrule
    mAP       & 40.1 & 41.3 & 41.6 & 41.8 & 42.2 & \textbf{42.3} & 42.1 \\
    AP$_{50}$ & 58.2 & 59.4 & 59.5 & 59.8 & \textbf{60.1} & 60.0 & 59.8 \\
    AP$_{75}$ & 43.1 & 44.7 & 44.9 & 45.4 & \textbf{45.9} & \textbf{45.9} & 45.8 \\
    % AP$_{S}$  & 23.3 & 23.9 & 24.6 & 25.2 & 25.3 & \textbf{25.4} & 24.4 \\
    % AP$_{M}$  & 44.4 & 45.2 & 45.3 & 45.9 & \textbf{46.4} & 46.3 & \textbf{46.4} \\
    % AP$_{L}$  & 52.5 & 53.1 & 54.3 & 54.3 & \textbf{55.0} & 54.3 & 54.6 \\
    
    \bottomrule
  \end{tabular}
  \end{center}
  \vspace{-1mm}
  \caption{Ablation study of hyper-parameter $\alpha_{2}$ on GFocal Res101-Res50. To show the sensitivity of $\mathcal{L}_{loc}^{dis}(x)$, we fix $\alpha_{1}=0$. Boldface indicates the best results.}
  \vspace{-3mm}
  \label{table:aplha2ablation}
\end{table}

\vspace{0.1cm}
\noindent \textbf{Self-KD.}
We have demonstrated the effectiveness of our proposed method for knowledge transfer from a strong teacher to a compact student in Table~\ref{table:main results}.
However, in cases where a stronger teacher is not available, self-KD~\cite{furlanello2018born,zhang2019your} can still be employed for classification tasks. 
We apply $S_{det} = T_{det}$ to the dense object detection task with our method, where $S_{det}$ is the student detector and $T_{det}$ is the teacher detector.
Table~\ref{table:selfkd} shows that our proposed method can still yield performance gains under the self-KD strategy.

\begin{table}[h]

  \begin{center}
\begin{tabular}{l|c|ccc}
    \toprule
    Method & Self-KD & mAP & AP$_{50}$ & AP$_{75}$\\
    \midrule
    \multirow{2}{*}{GFocal-Res50} &             & 40.1 & 58.2 & 43.1\\
                                  &  \checkmark & \textbf{40.9} & \textbf{59.1} & \textbf{44.2}\\
    \midrule
    \multirow{2}{*}{GFocal-Res34} &             & 38.9 & 56.6 & 42.2\\
                                  &  \checkmark & \textbf{39.4} & \textbf{57.2} & \textbf{42.6} \\
    \midrule
    \multirow{2}{*}{GFocal-Res18} &             & 35.8 & 53.1 & 38.2\\
                                  &  \checkmark & \textbf{36.2} & \textbf{53.5} & \textbf{38.9}\\
    
    \bottomrule
  \end{tabular}
  \end{center}
  \caption{Quantitative results of proposed method under the self-KD strategy. Boldface indicates the best results.}
  \vspace{-4mm}
  \label{table:selfkd}
\end{table}

\begin{figure}[!t]
  \centering
  \graphicspath{{images/}}
  \includegraphics[width=1.0\linewidth]{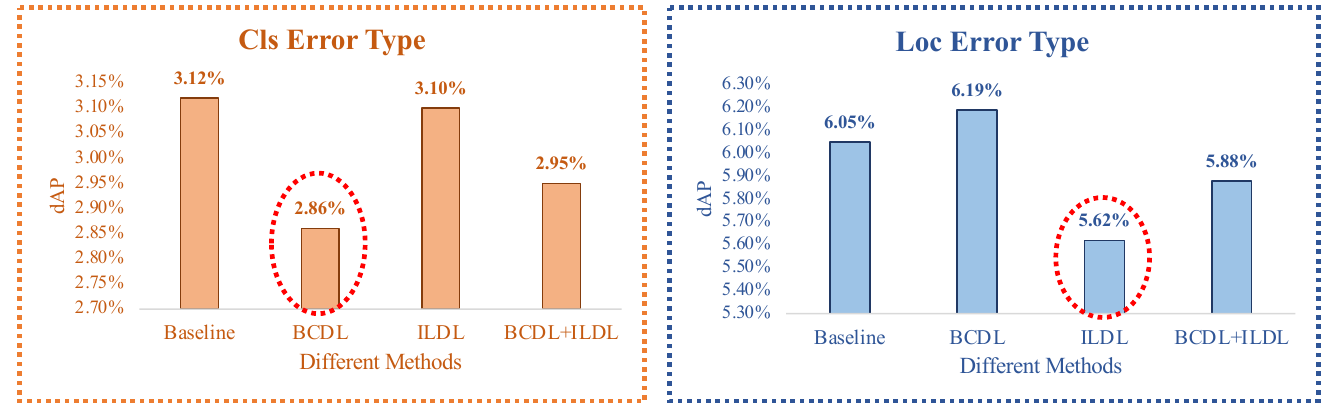}
  \caption{Error analysis conducted using the TIDE toolbox~\cite{bolya2020tide}. 
  The decrease in average precision (dAP) resulting from two types of errors (\emph{i.e.}, Cls, Loc)~\cite{bolya2020tide} is reported. 
  The student model without any distillation losses is denoted as ``Baseline", while the use of Binary Classification Distillation Loss and the application of IoU-based Localization Distillation Loss are denoted as ``BCDL" and ``ILDL", respectively.}
  \vspace{-6mm}
  \label{TIDE}
\end{figure}

\vspace{0.1cm}
\noindent \textbf{Error Analysis.}
The TIDE toolbox~\cite{bolya2020tide} is used to analyze the distribution of error types, as presented in Figure~\ref{TIDE}. 
The {\bf Cls} error type indicated correctly localized but misclassified predictions, and the {\bf Loc} error type indicated correctly classified but incorrectly localized predictions. 
The results showed two key findings: $(i)$ The Binary Classification Distillation Loss effectively reduced {\bf Cls} errors but did not contribute to reducing {\bf Loc} errors. 
$(ii)$ The IoU-based Localization Distillation Loss effectively reduced {\bf Loc} errors but did not contribute to reducing {\bf Cls} errors.
These results provide further evidence of the efficacy of Binary Classification Distillation Loss and IoU-based Localization Distillation Loss in enhancing classification and localization performance, respectively.

\section{Conclusion}
Our study reveals the cross-task protocol inconsistency is the reason behind the inefficiency of original classification distillation in dense object detection. 
To solve this problem, we present a novel Binary Classification Distillation Loss.
Besides, we design an IoU-based Localization Distillation Loss for eliminating the need for specific structure. 
Experimental results demonstrate the effectiveness of our proposed method, especially in improving classification distillation performance.
We expect that our work will provide valuable insights and encourage further research into logit-based distillation methods.

\paragraph{Acknowledgements.}
This work is supported in part by National Natural Science Foundation of China under Grant U20A20222, National Science Foundation for Distinguished Young Scholars under Grant 62225605, National Key Research and Development Program of China under Grant 2020AAA0107400, Hikvision Cooperation Fund, The Ng Teng Fong Charitable Foundation in the form of ZJU-SUTD IDEA Grant, 188170-11102, Zhejiang Key Research and Development Program under Grant 2023C03196, and sponsored by CCF-AFSG Research Fund, CAAI-HUAWEI MindSpore Open Fund as well as CCF-Zhipu AI Large Model Fund (CCF-Zhipu202302).

{\small
\bibliographystyle{ieee_fullname}
\bibliography{egbib}
}

\twocolumn[
\begin{@twocolumnfalse}
\section*{\centering{\Large{Appendix for Bridging Cross-task Protocol Inconsistency \\ for Distillation in Dense Object Detection\\[25pt]}}}
\end{@twocolumnfalse}
]

\appendix

% \begin{center}
% {\Large \bf Supplement Materials for Bridging Cross-task Protocol Inconsistency \\ for Distillation in Dense Object Detection\par}
% \end{center}

% In the supplementary material, we provide the following experimental results and details:

% \begin{itemize}

% \item{Section~\ref{algorithm}: Algorithm to calculate the distillation loss.}

% \item{Section~\ref{implement_all}: Implementation details.}

% \item{Section~\ref{experiment}: More experimental results about self-KD, heterogeneous backbones, base detectors, Pascal VOC dataset, inheriting initialization and response-based distillation.}

% \item{Section~\ref{visualization}: Visualization about intermediate training phases and predicted results.}

% \end{itemize}

%%%%%%%%% BODY TEXT
\section{Algorithm}
\label{algorithm}

In this section, we propose an algorithm to calculate the distillation loss, which is illustrated in Algorithm~\ref{algorithm1}.

\begin{algorithm}[htb]
\caption{The algorithm to calculate the distillation loss.}
\label{algorithm1}
\begin{algorithmic}[1]
\Require Training data ${x_i}_{\{i=1,\cdots,n\}}$, student dense object detector $S_{det}$, parameters $\theta_s$ of $S_{det}$, student dense object detector $T_{det}$, parameters $\theta_t$ of $T_{det}$, positions $pos$ of anchors
\State Uniformly sample a minibatch of training data $B^{(t)}$
\For{$x_i \in B^{(t)}$}
\State $l^s$, $o^s$ = $S_{det}(x_i;\theta_s)$
\State $l^t$, $o^t$ = $T_{det}(x_i;\theta_t)$
\Procedure{Classification}{$l^s$, $l^t$}
\State ${p^s}'$ = Sigmoid($l^s$)
\State ${p^t}'$ = Sigmoid($l^t$)
\State $w$ = $\left| {p^t}' - {p^s}' \right|$
\State $\mathcal{L}_{cls}^{dis}(x_i)$ = $\sum$ $w$ $\cdot$ $BCE$(${l^s}'$,${l^t}'$)
\EndProcedure
\Procedure{Localization}{$o^s$, $o^t$}
\State $b^s$ = Decoder($pos$, $o^s$)
\State $b^t$ = Decoder($pos$, $o^t$)
\State $u'$ = IoU($b^s$, $b^t$)
\State $\mathcal{L}_{loc}^{dis}(x_i)$ = $\sum_{i=1}^{n} \mathop{max}(w_{.,j}) \cdot (1 - u_{i}')$
\EndProcedure
\EndFor
\end{algorithmic}
\end{algorithm}

\section{Implementation Details}
\label{implement_all}
\subsection{Main Experiment}
\label{implement}
Our implementation is based on Pytorch and mmdetection~\cite{chen2019mmdetection}. 
Different training schedules are set to ensure fair comparison with previous methods, such as 1x (namely 12 epochs) and 2x (namely 24 epochs). 
We use SGD optimizer with momentum and weight decay set to 0.9 and 0.0001, respectively. 
The initial learning rate is set to 0.01. 
Our proposed method employs $\alpha_{1}$ and $\alpha_{2}$ to balance the classification and localization distillation losses, which are set to 1.0 and 4.0, respectively. 
All experiments are conducted on 8 RTX 3090 GPUs, with a batch size of 2 images per GPU.

\subsection{Combined with Feature-based Methods}
Our implementation is based on Pytorch and mmrazor~\cite{2021mmrazor}. 
We adopt the same training schedules, SGD optimizer, and learning rate settings as described in Section~\ref{implement}. 
The hyperparameters $\alpha_{1}$ and $\alpha_{2}$ are set to 0.25 and 2.0, respectively. 
All experiments are conducted on 8 RTX 3090 GPUs with 2 images per GPU.

\section{More Experiment}
\label{experiment}

\begin{table*}[h]
  \begin{center}
  
  \begin{tabular}{l|c|cccccc}
    \toprule
    Method & Schedule & mAP & AP$_{50}$ & AP$_{75}$ & AP$_{S}$ & AP$_{M}$ & AP$_{L}$\\

    \midrule
    \rowcolor{lightgray!45}GFocal-Res34(Student)   & 1x & 38.9 & 56.6 & 42.2 & 21.5 & 42.8 & 51.4\\
    LD~\cite{LD}  self-KD               & 1x & 38.6 & 56.0 & 41.7 & 21.0 & 42.4 & 50.4\\
    Ours self-KD     & 1x & \textbf{39.4} & \textbf{57.2} & \textbf{42.6} & \textbf{21.7} & \textbf{43.4} & \textbf{51.6}\\
    
    \midrule
    \rowcolor{lightgray!45}GFocal-Res18(Student)   & 1x & 35.8 & 53.1 & 38.2 & 18.9 & 38.9 & 47.9\\
    LD~\cite{LD} self-KD                & 1x & 35.0 & 52.1 & 37.7 & 18.6 & 38.6 & 46.0\\
    Ours Self-KD     & 1x & \textbf{36.2} & \textbf{53.5} & \textbf{38.9} & \textbf{19.3} & \textbf{39.6} & \textbf{48.3}\\
    
    \bottomrule
  \end{tabular}
  \end{center}
  \caption{Quantitative evaluation results of our proposed method and other logits-based distillation techniques for self-KD scenario on MS COCO \textit{val2017}.}
  \label{table:selfkd}
\end{table*}

\subsection{Self KD}
We have demonstrated the effectiveness of our proposed approach in transferring knowledge from a powerful teacher to a compact student.
Then, in cases where a stronger teacher model is not available, self-KD~\cite{furlanello2018born,zhang2019your} has emerged as a popular technique for classification. 
In the context of dense object detection, we simulate similar scenarios by setting $S_{det} = T_{det}$, where $S_{det}$ and $T_{det}$ denote the student and teacher detectors, respectively. 
Our approach also improves the performance under the self-KD strategy with lightweight detectors, as shown in Table~\ref{table:selfkd}. 
In contrast, LD~\cite{LD} leads to performance degradation in these scenarios.

\begin{table*}[h]
  \begin{center}
  
  \begin{tabular}{l|c|cccccc}
    \toprule
    Method & Schedule & mAP & AP$_{50}$ & AP$_{75}$ & AP$_{S}$ & AP$_{M}$ & AP$_{L}$\\
    \midrule
    \rowcolor{lightgray!45}GFocal-SwinT(Teacher)  & 2x & 47.3 & 66.2 & 51.4 & 31.8 & 50.9 & 60.7\\
    \rowcolor{lightgray!45}GFocal-Res50(Student)   & 1x & 40.1 & 58.2 & 43.1 & 23.3 & 44.4 & 52.5\\
    Ours     & 1x & 43.0 & 61.5 & 46.7 & 25.7 & 47.3 & 55.9\\

    \midrule
    \rowcolor{lightgray!45}GFocal-ResX101DCN(Teacher)   & 2x & 48.1 & 67.1 & 52.5 & 29.7 & 52.1 & 62.7\\
    \rowcolor{lightgray!45}GFocal-Res50(Student)   & 1x & 40.1 & 58.2 & 43.1 & 23.3 & 44.4 & 52.5\\
    Ours     & 1x & 42.6 & 61.4 & 46.4 & 26.1 & 46.4 & 55.1\\

    \midrule
    \rowcolor{lightgray!45}GFocal-Res50(Teacher)   & 2x & 42.9 & 61.2 & 46.5 & 27.3 & 46.9 & 53.3\\
    \rowcolor{lightgray!45}GFocal-Res50(Student)   & 1x & 40.1 & 58.2 & 43.1 & 23.3 & 44.4 & 52.5\\
    Ours     & 1x & 42.8 & 61.2 & 46.4 & 26.0 & 47.0 & 54.1\\

    \midrule
    \rowcolor{lightgray!45}GFocal-Res101(Teacher)  & 2x & 44.9 & 63.1 & 49.0 & 28.0 & 49.1 & 57.2\\
    \rowcolor{lightgray!45}GFocal-MobileNetv2(Student)   & 1x & 32.6 & 48.5 & 34.9 & 18.0 & 34.6 & 43.5\\
    Ours     & 1x & 35.1 & 51.8 & 37.8 & 19.1 & 37.9 & 45.6\\

    \bottomrule
  \end{tabular}
  \end{center}
  \caption{Quantitative evaluation results of proposed distillation method for heterogeneous backbones on MS COCO \textit{val2017}.}
  \label{table:heterogeneous}
\end{table*}

\subsection{Heterogeneous Backbone}
Recently, powerful backbones such as Swin-Transformer have exhibited remarkable performance in various computer vision tasks. 
Nonetheless, CNN-based dense object detectors remain extensively employed in practical applications due to their high speed and ease of deployment.
Due to the large gap in feature representations between the two architectures, applying feature-based methods from Transformer-based detectors to CNN-based detectors is challenging. 
To address this issue, we propose a prediction-level distillation method that is feature-free and particularly suitable for this task. 
As a result, we can use more powerful teacher detectors to enhance the performance of compact student detectors. 
As demonstrated in Table~\ref{table:heterogeneous}, our proposed method is effective when transferring knowledge between detectors with heterogeneous backbones.

\begin{table*}[h]
  \begin{center}
  
  \begin{tabular}{l|c|cccccc}
    \toprule
    Method & Schedule & mAP & AP$_{50}$ & AP$_{75}$ & AP$_{S}$ & AP$_{M}$ & AP$_{L}$\\
    \midrule
    \rowcolor{lightgray!45}ATSS-Res101(Teacher)  & 1x & 41.5 & 59.9 & 45.2 & 24.2 & 45.9 & 53.3\\
    \rowcolor{lightgray!45}ATSS-Res50(Student)   & 1x & 39.4 & 57.6 & 42.8 & 23.6 & 42.9 & 50.3\\
    Ours     & 1x & {\bf 41.4(+2.0)} & {\bf 59.9(+2.3)} &  {\bf 45.1(+2.3)} & {\bf 25.1(+1.5)} & 45.6(+2.7) & 53.5(+3.2)\\
    PKD     & 1x & 41.3(+1.9) & 59.2(+1.6) & 44.6(+1.8) & 24.1(+0.5) & 45.6(+2.7) & 53.9(+3.6)\\
    PKD + Ours     & 1x & {\bf 41.4(+2.0)} & 59.5(+1.9) & 44.8(+2.0) & 23.7(+0.1) & {\bf 45.7(+2.8)} & {\bf 54.1(+3.8)}\\
    
    \midrule
    \rowcolor{lightgray!45}YOLOX-s (Teacher)  & 1x & 40.3 & 59.1 & 43.4 & 23.5 & 44.5 & 53.1\\
    \rowcolor{lightgray!45}YOLOX-tiny (Student)   & 1x & 31.8 & 49.0 & 33.8 & 12.3 & 34.9 & 47.8\\
    Ours     & 1x & 34.2(+2.4) & 52.0(+3.0) & 36.4(+2.6) & 14.7(+2.4) & 39.1(+4.2) & 50.0(+2.2)\\

    \bottomrule
  \end{tabular}
  \end{center}
  \caption{Quantitative evaluation results of different distillation methods for ATSS on MS COCO \textit{val2017}.}
  \label{table:atss}
\end{table*}

\subsection{Base Detector}
In this subsection, we evaluate our proposed method on additional base detectors, such as ATSS~\cite{zhang2020atss} and YOLOX. 
The results in Table~\ref{table:atss} indicate that our method achieves comparable gains to the state-of-the-art feature-based methods. 
The significant improvement in detector mAP on ATSS~\cite{zhang2020atss} and YOLOX further confirms the robust generalization ability of our proposed method.

\begin{table}[h]
  \begin{center}
  \begin{tabular}{l|cccccc}
    \toprule
    Method & mAP & AP$_{50}$ & AP$_{75}$\\
    \midrule
    \rowcolor{lightgray!45}GFocal-Res50 (Teacher)   & 56.4 & 79.1 & 61.3\\
    \rowcolor{lightgray!45}GFocal-Res18 (Student)   & 52.2 & 75.8 & 56.4 \\
    Ours     & 55.2 & 78.1 & 59.2\\
    
    \bottomrule
  \end{tabular}
  \end{center}
  \caption{Quantitative evaluation results on Pascal VOC.}
  \label{table:voc}
\end{table}

\subsection{Pascal VOC Dataset}
Many recent object detection distillation methods, e.g., ~\cite{MGD,PKD}, only report experimental results on COCO.
We follow their experimental setups. 
Additionally, we expand our evaluations on Pascal VOC.
Table~\ref{table:voc} shows that our method increases the performance from 52.2 to 55.2.

\begin{table}[h]
  \begin{center}
  \begin{tabular}{l|cccccc}
    \toprule
    Method & mAP & AP$_{50}$ & AP$_{75}$\\
    \midrule
    \rowcolor{lightgray!45}FCOS-Res101 (Teacher)   & 40.8 & 60.0 & 44.0\\  
    \rowcolor{lightgray!45}FCOS-Res50 (Student)   & 36.6 & 56.0 & 38.8 \\
    PKD + Ours w/o inheriting    & 40.2 & 59.5 & 43.0\\
    PKD + Ours w/ inheriting    & 40.7 & 60.0 & 43.5 \\
    \bottomrule
  \end{tabular}
  \end{center}
  \caption{Quantitative evaluation results with inherit strategy.}
  \label{table:inherit}
\end{table}

\subsection{Inheriting Initialization}
As shown in Table~\ref{table:inherit}, equipping with the inheriting strategy, the performance of our method further increases by 0.5 mAP.

\begin{table}[h]
  \begin{center}
  
  \begin{tabular}{l|ccc}
    \toprule
    Method & mAP & AP$_{50}$ & AP$_{75}$\\
    \midrule
    \rowcolor{lightgray!45}Retina-Res101 (Teacher) & 38.1 & 58.3 & 40.9\\
    \rowcolor{lightgray!45}Retina-Res50 (Student) & 36.2 & 55.8 & 38.8\\
    Response-based Distillation~\cite{dai2021general}   & 37.9 & 57.8 &  41.1\\
    Ours   & {\bf 39.2} & {\bf 59.1} &  {\bf 42.4}\\
    \bottomrule
  \end{tabular}
  \end{center}
  \caption{Quantitative evaluation results of our method and Response-based Distillation in GID.}
  \label{table:atss}
\end{table}

\subsection{Response-based Distillation}
GID~\cite{dai2021general} proposes response-based distillation.
Our method is different with GID~\cite{dai2021general}.
On the one hand, the response-based distillation in GID is motivated by ``the definition of outputs from the detector head varies from model to model".
In contrast, our method is motivated by an in-depth analysis of the main challenge faced by logit-based distillation techniques in object detection.
We identify the cross-task protocol inconsistency between distillation and classification and propose cross-task consistent protocols as a solution.
This finding offers an interesting insight to the research community.
On the other hand, GID selects part regions of images (i.e., GIs) for distillation.
Our method incorporates all regions of the image in distillation, and we use score-aware weighting for different regions, eliminating the need for complex GI designs.
We compare our method with GID in Table~\ref{table:atss}. 
The results in Table~\ref{table:atss} indicate the superiority of our method over GID.

\section{Visualization}
\label{visualization}

\begin{figure*}[h]
  \centering
   \includegraphics[width=1.0\linewidth]{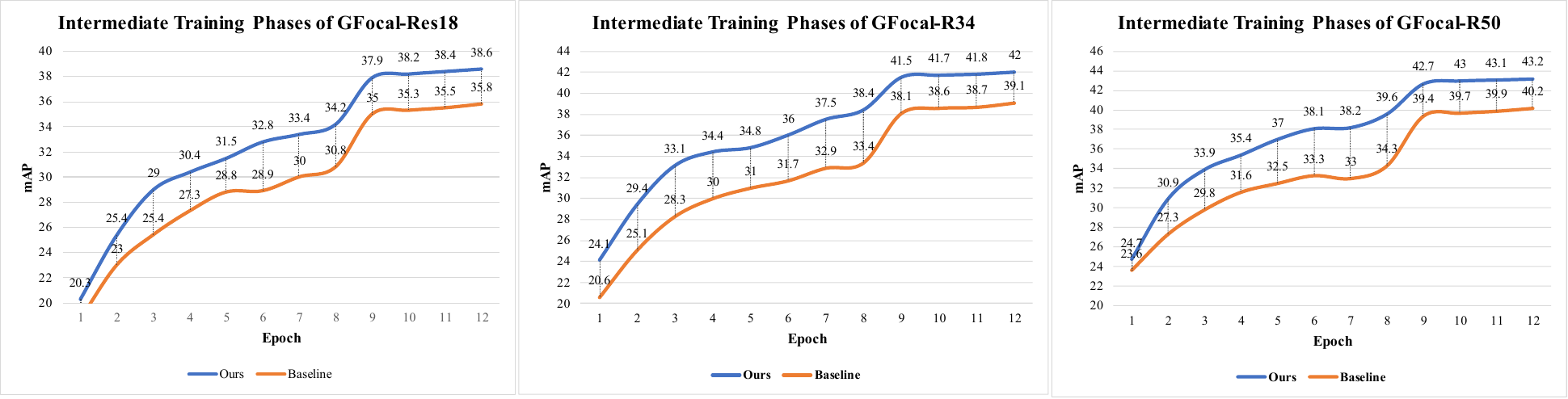}
   \caption{Visualization of intermediate training phases on GFocal-Res18, GFocal-R34 and GFocal-R50.}
   \label{fig:convergence}
\end{figure*}

\subsection{Intermediate Phases of Model Training}
Our approach improves the performance of the model at different intermediate stages of the training process, which correspond to different epochs, when compared to the baseline. 
As illustrated in Figure~\ref{fig:convergence}, by the 8th epoch, the performance gap between our method and the baseline is significantly larger than that at the final epoch. 
Specifically, for GFocal R34, our approach achieves a performance improvement of 5$\%$ at the 8th epoch, which is higher than the 2.9$\%$ improvement observed at the final epoch.
These results suggest that our method achieves faster convergence compared to the baseline.

\subsection{Prediction Visualization}
\begin{figure*}[h]
  \centering
   \includegraphics[width=0.95\linewidth]{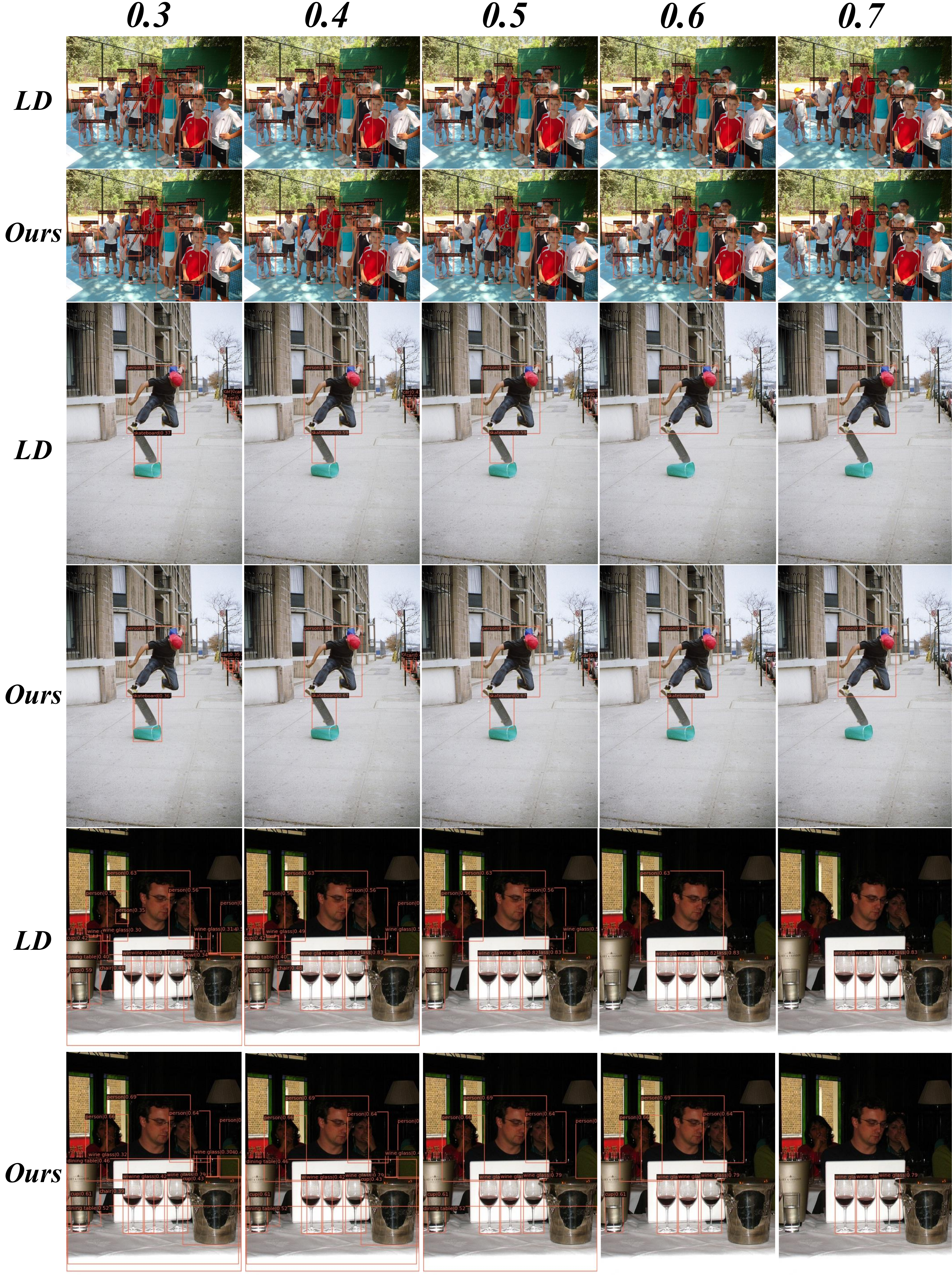}
   \caption{Visualization compared to LD~\cite{LD}. \textit{0.3} means under score threshold = 0.3. Best viewed in color.}
   \label{fig:visualization}
\end{figure*}

We provide visual evidence to validate the effectiveness of our proposed method by presenting images from MS COCO~\cite{coco} \textit{val2017} with various score thresholds. 
Figure~\ref{fig:visualization} depicts that our method outperforms LD~\cite{LD} in detecting more high-quality bounding boxes, particularly under high score thresholds. 
This result implies that our proposed method offers a more favorable classification score distribution.

\end{document}